%% file: Workshop_1.tex
\documentclass{workshop2002}
\usepackage{natbib}
\usepackage[dvips]{graphicx}
\usepackage{theorem}
\usepackage{subeqn}
\input{macros.tex}

\theoremstyle{plain}

\begin{document}
\title {Design of a Three-Axis Isotropic Parallel Manipulator for Machining Applications: \goodbreak The Orthoglide}
\author{Philippe Wenger, Damien Chablat
\affiliation{Institut de Recherche en Communications et
Cybern\'etique de Nantes (IRCCyN) \\
 1, rue de la No\"e, 44321 Nantes, France\\
 email: Philippe.Wenger@irccyn.ec-nantes.fr\\
 email: Damien.Chablat@irccyn.ec-nantes.fr}}
\maketitle
\begin{abstract}
The orthoglide is a 3-DOF parallel mechanism designed at IRCCyN
for machining applications. It features three fixed parallel
linear joints which are mounted orthogonally and a mobile platform
which moves in the Cartesian $x$-$y$-$z$ space with fixed
orientation. The orthoglide has been designed as function of a
prescribed Cartesian workspace with prescribed kinetostatic
performances. The interesting features of the orthoglide are a
regular Cartesian workspace shape, uniform performances in all
directions and good compactness. A small-scale prototype of the
orthoglide under development is presented at the end of this
paper.
\end{abstract}
\section{Introduction}
Parallel kinematic machines (PKM) are interesting
alternative designs for high-speed machining applications and have been
 attracting the interest
of more and more researchers and companies. Since the first
prototype presented in 1994 during the IMTS in Chicago by
Gidding\&Lewis (the Variax), many other prototypes have appeared.

However, the existing PKM suffer from two major drawbacks, namely,
a complex Cartesian workspace and highly non linear input/output
relations. For most PKM, the Jacobian matrix which relates the
joint rates to the output velocities is not constant and not
isotropic. Consequently, the performances (e.g. maximum speeds,
forces accuracy and rigidity) vary considerably for different
points in the Cartesian workspace and for different directions at
one given point. This is a serious drawback for machining
applications (\cite{Kim:1997,Treib:1998,Wenger:1999b}). To be of
interest for machining applications, a PKM should preserve good
workspace properties, that is, regular shape and acceptable
kinetostatic performances throughout. In milling applications, the
machining conditions must remain constant along the whole tool
path (\cite{Rehsteiner:1999a,Rehsteiner:1999b}). In many research
papers, this criterion is not taking into account in the
algorithmic methods used for the optimization of the workspace
volume (\cite{Luh:1996,Merlet:1999}).

The orthoglide optimization is conducted to define a $3$-axis PKM
with the advantages a classical serial PPP machine tool but not
its drawbacks. Most industrial 3-axis machine-tool have a serial
PPP kinematic architecture with orthogonal linear joint axes along
the x, y and z directions. Thus, the motion of the tool in any of
these directions is linearly related to the motion of one of the
three actuated axes. Also, the performances are constant in the
most part of the Cartesian workspace, which is a parallelepiped.
The main drawback is inherent to the serial arrangement of the
links, namely, poor dynamic performances.

The orthoglide is a PKM with three fixed linear joints mounted
orthogonally. The mobile platform is connected to the linear
joints by three articulated parallelograms and moves in the
Cartesian x-y-z space with fixed orientation. Its workspace shape
is close to a cube whose sides are parallel to the planes $xy$,
$yz$ and $xz$ respectively. The optimization is conducted on the
basis of the size of a prescribed cubic workspace with bounded
velocity and force transmission factors. Two criteria are used for
the architecture optimization of the orthoglide, (i) the
conditioning of the Jacobian matrix of the PKM
(\cite{Golub:1989,Salisbury:1982,Angeles:1997}) and (ii) the
manipulability ellipsoid (\cite{Yoshikawa:1985}).

The first criterion leads to an isotropic architecture and to
homogeneous performances in the workspace. The second criterion
permits to optimize the actuated joint limits and the link lengths
of the orthoglide with respect to the aforementioned two criteria.

Next section presents the orthoglide. The kinematic equations and
the singularity analysis is detailed in Section~3. Section~4 is
devoted to the optimization process of the orthoglide and to the
presentation of the prototype.
\section{Description of the Orthoglide}
Most existing PKM can be classified into two main families. The
PKM of the first family have fixed foot points and variable length
struts and are generally called ``hexapods''. They have a
Stewart-Gought parallel kinematic architecture. Many prototypes
and commercial hexapod PKM already exist like the
Variax-Hexacenter (Gidding\&Lewis), the CMW300 (Compagnie
M\'ecanique des Vosges), the TORNADO 2000 (Hexel), the MIKROMAT 6X
(Mikromat/IWU), the hexapod OKUMA (Okuma), the hexapod G500
(GEODETIC). In this first family, we find also hybrid
architectures with a 2-axis wrist mounted in series to a 3-DOF
tripod positioning structure (the TRICEPT from Neos Robotics).

The second family of PKM has been more recently investigated. In
this category we find the HEXAGLIDE (ETH Z\"{u}rich) which
features six parallel (also in the geometrical sense) and coplanar
linear joints. The HexaM (Toyoda) is another example with non
coplanar linear joints. A 3-axis translational version of the
hexaglide is the TRIGLIDE (Mikron), which has three coplanar and
parallel linear joints. Another 3-axis translational PKM is
proposed by the ISW Uni Stuttgart with the LINAPOD. This PKM has
three vertical (non coplanar) linear joints. The URANE SX (Renault
Automation) and the QUICKSTEP (Krause \& Mauser) are 3-axis PKM
with three non coplanar horizontal linear joints. The SPRINT Z3
(DS Technology) is a 3-axis PKM with one degree of translation and
two degrees of rotations. A hybrid parallel/serial PKM with three
parallel inclined linear joints and a two-axis wrist is the GEORGE
V (IFW Uni Hanover).

PKMs of the second family are more interesting because the
actuators are fixed and thus the moving masses are lower than in
the hexapods and tripods.

\begin{figure}[!ht]
  \begin{center}
    \includegraphics[width=67mm,height=47mm]{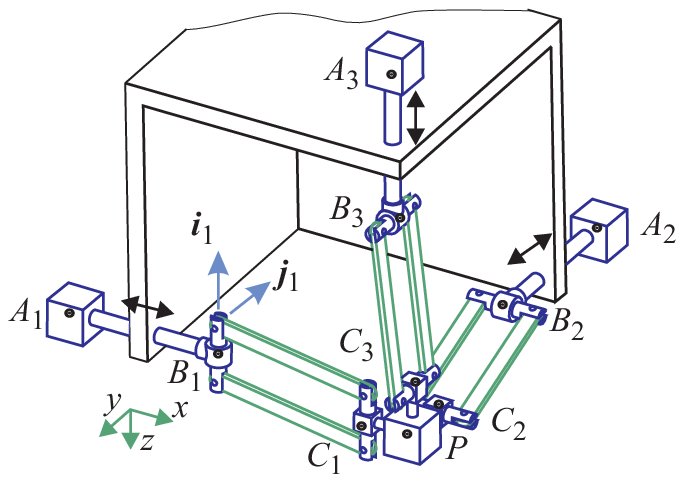}
    \caption{Orthoglide kinematic architecture}
    \protect\label{figure:Orthoglide}
           \includegraphics[width=48mm,height=38mm]{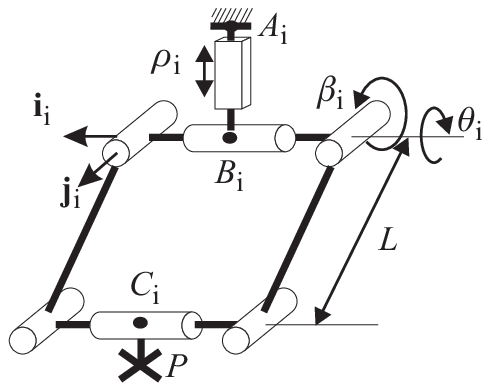}
           \caption{Leg kinematics}
           \protect\label{figure:Leg_Kinematic}
    \end{center}
\end{figure}

The  orthoglide presented in this article is a $3$-axis
translational parallel kinematic machine with variable foot points
and fixed length struts. Figure~\ref{figure:Orthoglide} shows the
general kinematic architecture of the orthoglide.

The orthoglide has three parallel $PRPaR$ identical chains (where
$P$, $R$ and $Pa$ stands for Prismatic, Revolute and Parallelogram
joint, respectively). The actuated joints are the three orthogonal
linear joints. These joints can be actuated by means of linear
motors or by conventional rotary motors with ball screws. The
output body is connected to the linear joints through a set of
three parallelograms of equal lengths $L~=~B_iC_i$, so that it can
move only in translation. The first linear joint axis is parallel
to the $x$-axis, the second one is parallel to the $y$-axis and
the third one is parallel to the $z$-axis. In
figure~\ref{figure:Orthoglide}, the base points $A_1$, $A_2$ and
$A_3$ are fixed on the $i^{th}$ linear axis such that
$A_1A_2=~A_1A_3=~A_2A_3$, $B_i$ is at the intersection of the
first revolute axis $\negr i_i$ and the second revolute axis
$\negr j_i$ of the $i^{th}$ parallelogram, and $C_i$ is at the
intersection of the last two revolute joints of the $i^{th}$
parallelogram. When each $B_iC_i$ is aligned with the linear joint
axis $A_iB_i$ , the orthoglide is in an isotropic configuration
(see ~\ref{Section4}) and the tool center point $P$ is located at
the intersection of the three linear joint axes. In this
configuration, the base points $A_1$, $A_2$ and $A_3$ are equally
distant from $P$. The symmetric design and the simplicity of the
kinematic chains (all joints have only one degree of freedom,
Fig.~\ref{figure:Leg_Kinematic}) should contribute to lower the
manufacturing cost of the orthoglide.

The orthoglide is free of singularities and self-collisions. The
workspace has a regular, quasi-cubic shape. The input/output
equations are simple and the velocity transmission factors are
equal to one along the $x$, $y$ and $z$ direction at the isotropic
configuration, like in a serial $PPP$ machine
(\cite{Wenger:2000}).
\section{Kinematic Equations and Singularity Analysis}
\subsection{Static Equations}
Let $\theta_i$ and $\beta_i$ denote the joint angles of the
parallelogram about the axes $\negr i_i$ and $\negr j_i$,
respectively (fig.~\ref{figure:Leg_Kinematic}). Let $\negr
\rho_1$, $\negr \rho_2$, $\negr \rho_3$ denote the linear joint
variables, $\negr \rho_i=A_iB_i$. In a reference frame (O, $x$,
$y$, $z$) centered at the intersection of the three linear joint
axes (note that the reference frame has been translated in
Fig.~\ref{figure:Orthoglide} for more legibility) , the position
vector \negr p of the tool center point $P$ can be defined in
three different ways:
 \bseq
 \beqa
   \negr p\!\!&=&\!\!
     \left[\begin{array}{c}
             a + \rho_1 + \cos(\theta_1) \cos(\beta_1) L + e \\
             \sin(\theta_1) \cos(\beta_1) L \\
             - \sin(\beta_1) L
         \end{array}
   \right] \\
   \negr p\!\!&=&\!\!
     \left[\begin{array}{c}
             -\sin(\beta_2) L \\
             a + \rho_2 + \cos(\theta_2) \cos(\beta_2) L + e \\
             \sin(\theta_2) \cos(\beta_2) L \\
         \end{array}
   \right] \\
   \negr p\!\!&=&\!\!
     \left[\begin{array}{c}
             \sin(\theta_3) \cos(\beta_3) L \\
             - \sin(\beta_3) L \\
             a + \rho_3 + \cos(\theta_3) \cos(\beta_3) L + e \\
         \end{array}
   \right]
 \eeqa
 \label{e:p-poistion}
 \eseq
where $a=OA_i$, $e=C_iP$ and we recall that $L=B_iC_i$, $\negr
\rho_i=A_iB_i$.
\subsection{Kinematic Equations}
Let $\dot{\gbf {\rho}}$ be referred to as the vector of actuated
joint rates and $\dot{\negr p}$ as the velocity vector of point
$P$:
 \bed
    \dot{\gbf{\rho}}=
    [\dot{\rho}_1~\dot{\rho}_2~\dot{\rho}_3]^T
   ,\quad
    \dot{\negr p}=
    [\dot{x}~\dot{y}~\dot{z}]^T
 \eed
$\dot{\negr p}$ can be written in three different ways by
traversing the three chains $A_iB_iC_iP$:
 \bseq
 \beqa
    \dot{\negr p} \!\!\!&=&\!\!\!\!
    \negr n_1 \dot{\rho}_1 +
    (\dot{\theta}_1 \negr i_1 + \dot{\beta}_1 \negr j_1)
    \times
    (\negr c_1 - \negr b_1) \\
    \dot{\negr p} \!\!\!&=&\!\!\!\!
    \negr n_2 \dot{\rho}_1 +
    (\dot{\theta}_2 \negr i_2 + \dot{\beta}_2 \negr j_2)
    \times
    (\negr c_2 - \negr b_2) \\
    \dot{\negr p} \!\!\!&=&\!\!\!\!
    \negr n_3 \dot{\rho}_3 +
    (\dot{\theta}_3 \negr i_3 + \dot{\beta}_3 \negr j_3)
    \times
    (\negr c_3 - \negr b_3)
 \eeqa
 \label{equation:cinematique}
 \eseq
where $\negr b_i$ and $\negr c_i$ are the position vectors of the
points $B_i$ and $C_i$, respectively, and $\negr n_i$ is the
direction vector of the linear joints, for $i=1, 2, $3.
\subsection{Singular configurations}
We want to eliminate the two idle joint rates $\dot{\theta}_i$ and
$\dot{\beta}_i$ from Eqs.~(\ref{equation:cinematique}a--c), which
we do upon dot-multiplying Eqs.~(\ref{equation:cinematique}a--c)
by $\negr c_i - \negr b_i$:
 \bseq
 \beqa
   (\negr c_1 - \negr b_1)^T \dot{\negr p} &=&
   (\negr c_1 - \negr b_1)^T
   \negr n_1 \dot{\rho}_1 \\
   (\negr c_2 - \negr b_2)^T \dot{\negr p} &=&
   (\negr c_2 - \negr b_2)^T
   \negr n_2 \dot{\rho}_2 \\
   (\negr c_3 - \negr b_3)^T \dot{\negr p} &=&
   (\negr c_3 - \negr b_3)^T
   \negr n_3\dot{\rho}_3
 \eeqa
 \label{equation:cinematique-2}
 \eseq
Equations (\ref{equation:cinematique-2}a--c) can now be cast in
vector form, namely
 \bed
   \negr A \dot{\bf p} = \negr B \dot{\gbf \rho}
 \eed
where \negr A and \negr B are the parallel and serial Jacobian
matrices, respectively:
 \bseq
 \beqa
   \negr A =
   \left[\begin{array}{c}
           (\negr c_1 - \negr b_1)^T \\
           (\negr c_2 - \negr b_2)^T \\
           (\negr c_3 - \negr b_3)^T
         \end{array}
   \right] \\
   \negr B =
   \left[\begin{array}{ccc}
            \eta_1&
            0 &
            0 \\
            0 &
            \eta_2&
            0 \\
            0 &
            0 &
            \eta_3
         \end{array}
   \right]
 \eeqa
 \label{equation:A_et_B}
 \eseq
with $\eta_i= (\negr c_i - \negr b_i)^T \negr n_i $ for $i =
1,2,3$.

\begin{figure}[!ht]
    \begin{center}
           \includegraphics[width=82mm,height=68mm]{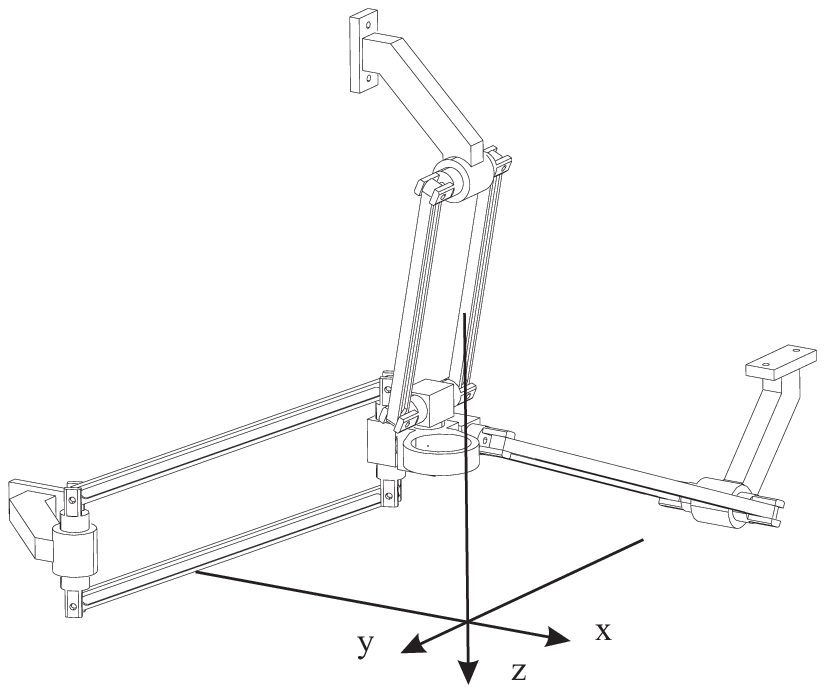}
           \caption{Parallel singular configuration when $B_iC_i$ are coplanar}
           \protect\label{figure:Coplanarsing}
           \includegraphics[width=45mm,height=56mm]{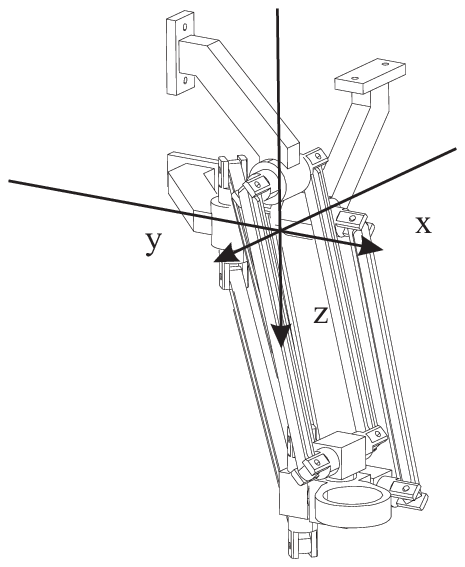}
           \caption{Parallel singular configuration when $B_iC_i$ are parallel}
           \protect\label{figure:Parallelsing}
    \end{center}
\end{figure}

The parallel singularities (\cite{Chablat:1998}) occur when the
determinant of the matrix \negr A vanishes, {\it i.e.} when
$det(\negr A)=0$. In such configurations, it is possible to move
locally the mobile platform whereas the actuated joints are
locked. These singularities are particularly undesirable because
the structure cannot resist any force.
Eq.~(\ref{equation:A_et_B}a) shows that the parallel singularities
occur when:
 \bed
    (\negr c_1 - \negr b_1) =
    \alpha  (\negr c_2 - \negr b_2) +
    \lambda (\negr c_3 - \negr b_3)
 \eed
that is when the points $B_1$, $C_1$, $B_2$, $C_2$, $B_3$ and
$C_3$ are coplanar (Fig.~\ref{figure:Coplanarsing}). A particular
case occurs when the links $B_iC_i$ are parallel
(Fig.~\ref{figure:Parallelsing}):
 \beqa
    (\negr c_1 - \negr b_1) &//&
    (\negr c_2 - \negr b_2)
    \quad {\rm and} \nonumber \\
    (\negr c_2 - \negr b_2) &//&
    (\negr c_3 - \negr b_3)
    \quad {\rm and} \nonumber \\
    (\negr c_3 - \negr b_3) &//&
    (\negr c_1 - \negr b_1) \nonumber
 \eeqa

Serial singularities arise when the serial Jacobian matrix \negr B
is no longer invertible {\it i.e.} when $det(\negr B)=0$. At a
serial singularity a direction exists along which any cartesian
velocity cannot be produced. Eq.~(\ref{equation:A_et_B}b) shows
that $\det(\negr B)=0$ when for one leg $i$, $(\negr b_i - \negr
a_i) \perp (\negr c_i - \negr b_i)$.

The optimization of the orthoglide will put the serial and
parallel singularities far away from the workspace (see
~\ref{Section4}).
\section{Design and Performance Analysis of the Orthoglide}
For usual machine tools, the Cartesian workspace is generally
given as a function of the size of a right-angled parallelepiped.
Due to the symmetrical architecture of the orthoglide, the
Cartesian workspace has a fairly regular shape in which it is
possible to include a cube whose sides are parallel to the planes
$xy$, $yz$ and $xz$ respectively
(Fig.~\ref{figure:Orthoglide_Workspace}).

The aim of this section is to define the dimensions of the
orthoglide as a function of the size $L_{Workspace}$ of a
prescribed cubic workspace with bounded transmission factors. We
first show that the orthogonal arrangement of the linear joints is
justified by the condition on the isotropy and manipulability: we
want the orthoglide to have an isotropic configuration with
velocity and force transmission factors equal to one. Then, we
impose that the transmission factors remain under prescribed
bounds throughout the prescribed workspace and we deduce the link
dimensions and the joint limits.
\subsection{Condition Number and Isotropic Configuration}
The Jacobian matrix is said to be isotropic when its condition
number attains its minimum value of one (\cite{Angeles:1997}). The
condition number of the Jacobian matrix is an interesting
performance index which characterises the distortion of a unit
ball under the transformation represented by the Jacobian matrix.
The Jacobian matrix of a manipulator is used to relate (i) the
joint rates and the Cartesian velocities, and (ii) the static load
on the output link and the joint torques or forces. Thus, the
condition number of the Jacobian matrix can be used to measure the
uniformity of the distribution of the tool velocities and forces
in the Cartesian workspace.
\subsection{Isotropic Configuration of the Orthoglide}
For parallel manipulators, it is more convenient to study the
conditioning of the Jacobian matrix that is related to the inverse
transformation, $\negr J^{-1}$. When \negr B is not singular,
$\negr J^{-1}$ is defined by:
 \bed
   \dot{\gbf \rho} = \negr J^{-1} \dot{\negr p}
   {\rm ~~with~~}
   \negr J^{-1} = \negr B^{-1} \negr A
 \eed
 Thus:
 \beqa
   \negr J^{-1} =
   \left[\begin{array}{c}
            (1/\eta_1) (\negr c_1 - \negr b_1)^T \\
            (1/\eta_2) (\negr c_2 - \negr b_2)^T \\
            (1/\eta_3) (\negr c_3 - \negr b_3)^T
         \end{array}
   \right]
   \label{equation:J}
 \eeqa
with $\eta_i= (\negr c_i - \negr b_i)^T \negr n_i $ for $i =
1,2,3$.

\begin{figure}[!ht]
    \begin{center}
           \includegraphics[width=82mm,height=83mm]{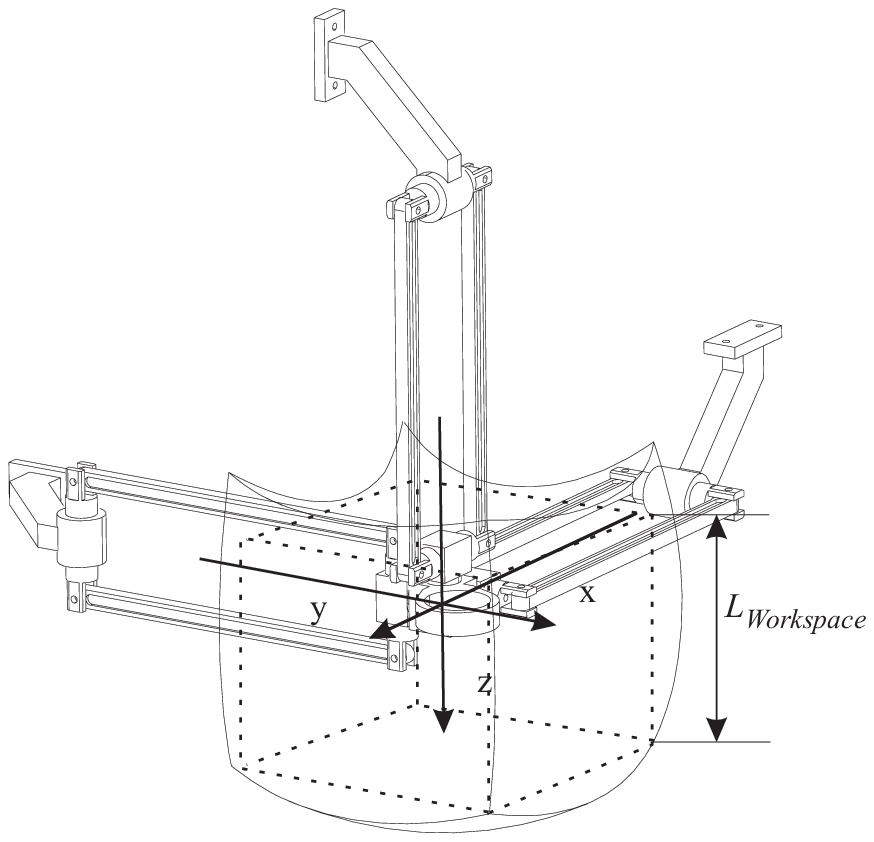}
           \caption{Cartesian workspace}
           \protect\label{figure:Orthoglide_Workspace}
     \includegraphics[width=78mm,height=73mm]{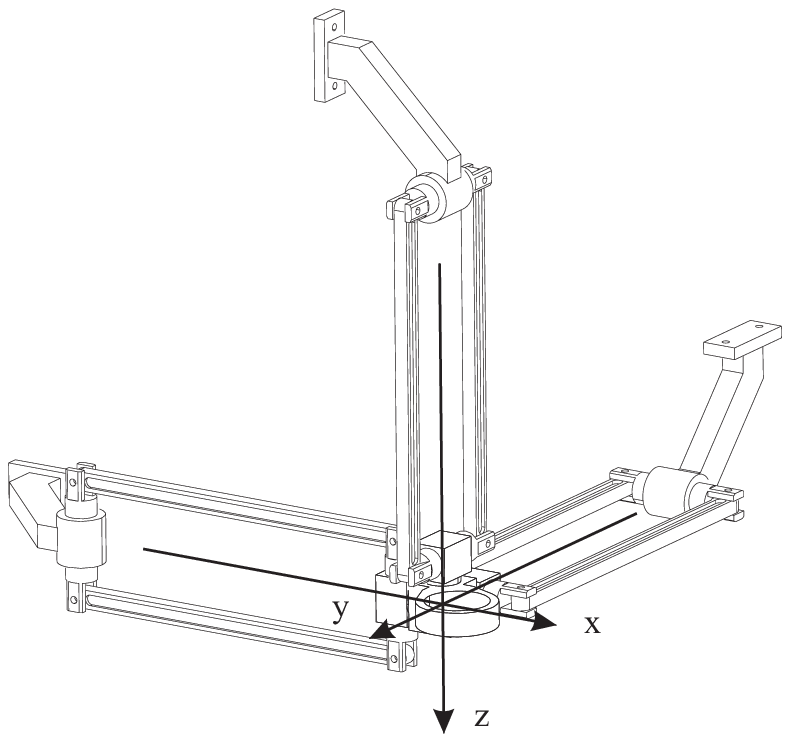}
     \caption{Isotropic configuration of the Orthoglide mecanism}
     \protect\label{figure:Isotropic_Configuration}
   \end{center}
 \end{figure}

The matrix $\negr J^{-1}$ is isotropic when $\negr J^{-1}\negr
J^{-T}=\sigma^2 \negr 1_{3 \times 3}$, where $\negr 1_{3 \times
3}$ is the $3 \times 3$ identity matrix. Thus, we must have,
 \bseq
   \be
     \frac{1}{\eta_1} ||\negr c_1 - \negr b_1|| =
     \frac{1}{\eta_2} ||\negr c_2 - \negr b_2|| =
     \frac{1}{\eta_3} ||\negr c_3 - \negr b_3||
   \ee
   \be
     (\negr c_1 - \negr b_1)^T
     (\negr c_2 - \negr b_2) = 0
   \ee
   \be
     (\negr c_2 - \negr b_2)^T
     (\negr c_3 - \negr b_3) = 0
   \ee
   \be
     (\negr c_3 - \negr b_3)^T
     (\negr c_1 - \negr b_1) = 0
   \ee
   \label{equation:isotropie}
 \eseq
Equation~(\ref{equation:isotropie}a) states that the orientation
between the axis of the linear joint and the link $B_iC_i$ must be
the same for each leg $i$.
Equations~(\ref{equation:isotropie}b--d) mean that the links
$B_iC_i$ must be orthogonal to each other.
Figure~\ref{figure:Isotropic_Configuration} shows the isotropic
configuration of the orthoglide. Note that the orthogonal
arrangement of the linear joints is not a consequence of the
isotropy condition, but it stems from the condition on the
transmission factors at the isotropic configuration (see next
section).
\subsection{Manipulability Analysis}
\begin{figure}[!ht]
    \begin{center}
        \includegraphics[width=49mm,height=36mm]{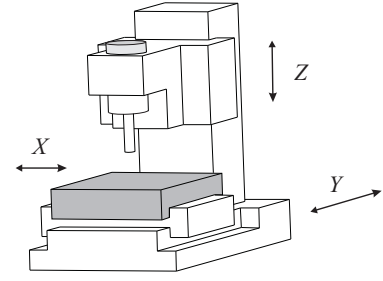}
        \caption{Typical industrial $3$-axis $PPP$ machine-tool}
        \protect\label{figure:machine_PPP}
    \end{center}
 \end{figure}

For a serial $PPP$ machine tool, Fig.~\ref{figure:machine_PPP}, a
motion of an actuated joint yields the same motion of the tool
(the transmission factors are equal to one). In the purpose on our
study, this factor is calculated from linear joint to the
end-effector.

For a parallel machine, these motions are generally not
equivalent. When the mechanism is close to a parallel singularity,
a small joint rate can generate a large velocity of the tool. This
means that the positioning accuracy of the tool is lower in some
directions for some configurations close to parallel singularities
because the encoder resolution is amplified. In addition, a
velocity amplification in one direction is equivalent to a loss of
rigidity in this direction.

The manipulability ellipsoids of the Jacobian matrix of robotic
manipulators was defined several years ago
(\cite{Salisbury:1982}). This concept has then been applied as a
performance index to parallel manipulators (\cite{Kim:1997}). Note
that, although the concept of manipulability is close to the
concept of condition number, these two concepts do not provide the
same information. The condition number quantifies the proximity to
an isotropic configuration, {\it i.e.} where the manipulability
ellipsoid is a sphere, or, in other words, where the transmission
factors are the same in all the directions, but it does not inform
about the value of the transmission factor.

The manipulability ellipsoid of $\negr J^{-1}$ is used here for
(i) justifying the orthogonal orientation of the linear joints and
(ii) defining the joint limits of the orthoglide such that the
transmission factors are bounded in the prescribed workspace.

We want the transmission factors to be equal to one at the
isotropic configuration like for a $PPP$ machine tool. This
condition implies that the three terms of
Eq.~(\ref{equation:isotropie}) must be equal to one:
   \beqa
     \frac{1}{\eta_1} ||\negr c_1 - \negr b_1|| =
     \frac{1}{\eta_2} ||\negr c_2 - \negr b_2|| =
     \frac{1}{\eta_3} ||\negr c_3 - \negr b_3|| = 1
     \label{eqution:isotropie-1}
   \eeqa
which implies that $(\negr b_i - \negr a_i)$ and  $(\negr c_i -
\negr b_i)$ must be collinear for each i.

Since, at this isotropic configuration, links $B_iC_i$ are
orthogonal, Eq.~(\ref{eqution:isotropie-1}) implies that the links
$A_iB_i$ are orthogonal, {\it i.e.} the linear joints are
orthogonal.
For joint rates belonging to a unit ball, namely,
$||\dot{\gbf \rho}|| \leq 1$, the Cartesian velocities belong to
an ellipsoid such that:
 \bed
   \dot{\gbf p}^T (\negr J \negr J^T) \dot{\gbf p} \leq 1
 \eed
The eigenvectors of matrix $(\negr J \negr J^T)^{-1}$ define the
direction of its principal axes of this ellipsoid and the square
roots $\xi_1$, $\xi_2$ and $\xi_3$ of the eigenvalues of $(\negr J
\negr J^T)^{-1}$ are the lengths of the aforementioned principal
axes. The velocity transmission factors in the directions of the
principal axes are defined by $\psi_1 = 1 / \xi_1$, $\psi_2 = 1 /
\xi_2$ and $\psi_3 = 1 / \xi_3$. To limit the variations of this
factor in the Cartesian workspace, we impose
 \be
   \psi_{min} \leq \psi_i \leq \psi_{max} 
   \label{e:velocity_limits}
 \ee
throughout the workspace. This condition determines the link
lengths and the linear joint limits. To simplify the problem, we
set $\psi_{min}=1/ \psi_{max}$.
\subsection{Design of the Orthoglide for a Prescribed Workspace}
\label{Section4}
The aim of this section is to define the position of the fixed
point $A_i$, the link lengths $L$ and the linear actuator range
$\Delta \rho$ with respect to the limits on the transmission
factors defined in Eq.~(\ref{e:velocity_limits}) and as a function
of the size of the prescribed workspace $L_{Workspace}$.

Our process of optimization is divided into three steps.
\begin{enumerate}
 \item
First, we determine two points $Q_1$ and $Q_2$ in the prescribed
cubic workspace such that if the transmission factor bounds are
satisfied at these points, they are satisfied in all the
prescribed workspace.
 \item
The points $Q_1$ and $Q_2$ are used to define the leg length $L$
as function of the size of the prescribed cubic workspace.
 \item
Finally, the positions of the base points $A_i$ and the linear
actuator range $\Delta \rho $ are calculated such that the
prescribed cubic workspace is fully included in the Cartesian
workspace of the orthoglide.
\end{enumerate}

\begin{figure}[!ht]
    \begin{center}
           \includegraphics[width=49mm,height=42mm]{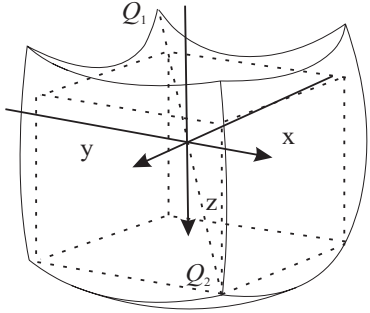}
           \caption{Points $Q_1$ and $Q_2$}
           \protect\label{figure:Workspace_Singularities}
    \end{center}
\end{figure}

\begin{figure}[!ht]
    \begin{center}
           \includegraphics[width=78mm,height=81mm]{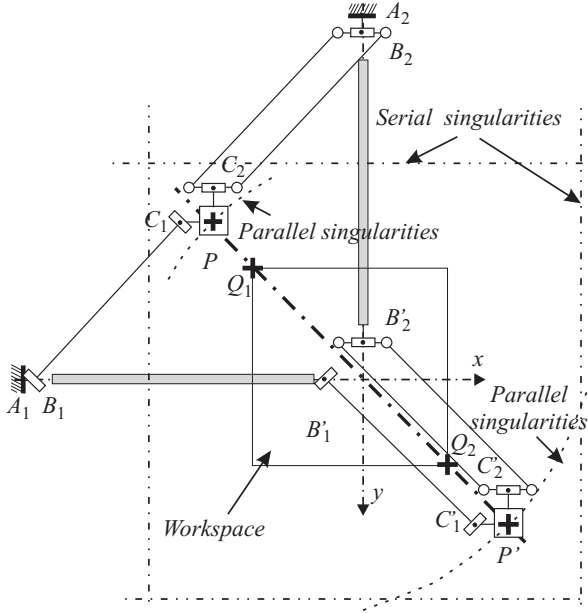}
           \caption{Points $Q_1$ and $Q_2$ and the singular configurations (top view)}
           \protect\label{figure:Workspace_Singularities2}
    \end{center}
\end{figure}

{\bf Step~1:} The transmission factors are equal to one at the
isotropic configuration. These factors increase or decrease when
the tool center point moves away from the isotropic configuration
and they tend towards zero or infinity in the vicinity of the
singularity surfaces. It turns out that the points $Q_1$ and $Q_2$
defined at the intersection of the workspace boundary with the
axis $x = y = z$ (figure \ref{figure:Workspace_Singularities}) are
the closest ones to the singularity surfaces, as illustrated in
figure \ref{figure:Workspace_Singularities2} which shows on the
same top view the orthoglide in the two parallel singular
configurations of figures ~\ref{figure:Coplanarsing} and
~\ref{figure:Parallelsing}. Thus, we may postulate the intuitive
result that if the prescribed bounds on the transmission factors
are satisfied at $Q_1$ and $Q_2$, then these bounds are satisfied
throughout the prescribed cubic workspace. Although we could not
derive a simple formal proof, we have verified numerically that
this result holds.

\begin{figure}[!ht]
    \begin{center}
           \includegraphics[width=71mm,height=68mm]{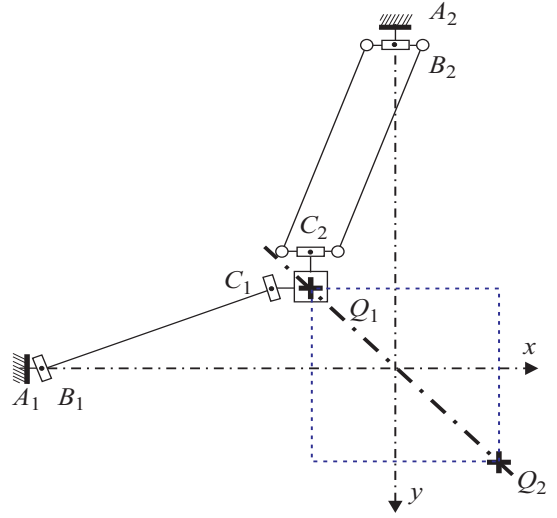}
           \caption{$Q_1$ configuration}
           \protect\label{figure:Workspace1}
    \end{center}
 \end{figure}

\begin{figure}[!ht]
    \begin{center}
           \includegraphics[width=73mm,height=68mm]{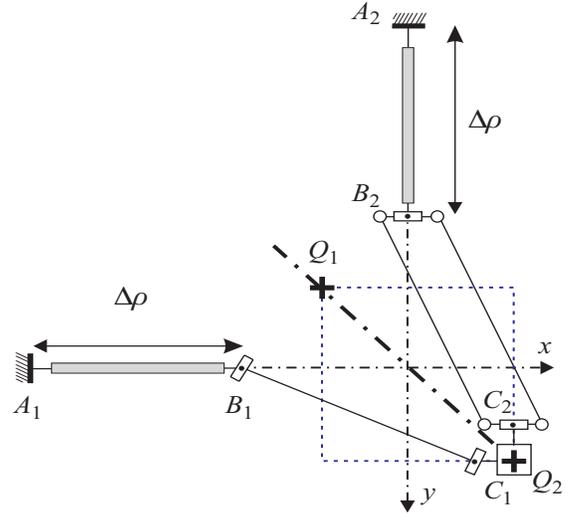}
           \caption{$Q_2$ configuration}
           \protect\label{figure:Workspace2}
    \end{center}
 \end{figure}

{\bf Step~2:} At the isotropic configuration, the angles
$\theta_i$ and $\beta_i$ are equal to zero by definition. When the
tool center point $P$ is at $Q_1$,
$\rho_1=\rho_2=\rho_3=\rho_{min}$ (Fig.~\ref{figure:Workspace1}).
When $P$ is at $Q_2$, $\rho_1=\rho_2=\rho_3=\rho_{max}$
(Fig.~\ref{figure:Workspace2}).

We pose $\rho_{min}=0$ for more simplicity.

\begin{figure}[!ht]
    \begin{center}
           \includegraphics[width=75mm,height=50mm]{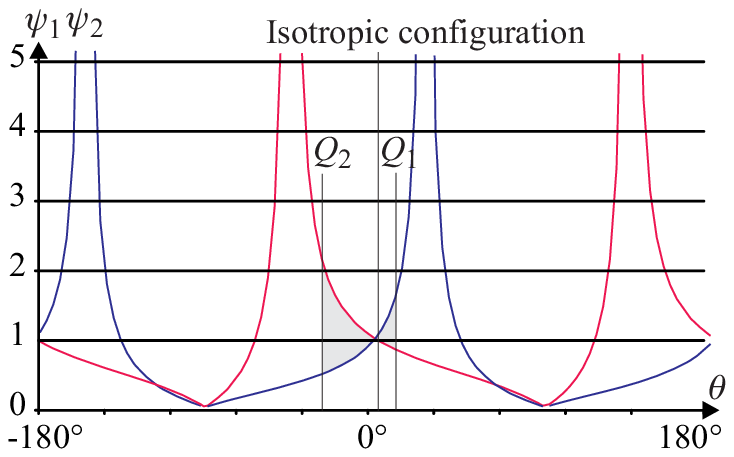}
           \caption{The three velocity transmission factors as
                    function of $\theta$ along the axis $(Q_1Q_2)$}
           \protect\label{figure:Variation_Xi}
    \end{center}
\end{figure}

On the axis $(Q_1Q_2)$, $\beta_1 = \beta_2 = \beta_3$ and
$\theta_1= \theta_2= \theta_3$. We note,
 \beqa
   \beta_1 = \beta_2 = \beta_3 = \beta \quad
   {\rm and} \quad
   \theta_1= \theta_2= \theta_3= \theta
   \label{e:beta_123}
 \eeqa
Upon substitution of Eq.~(\ref{e:beta_123}) into
Eqs.~(\ref{e:p-poistion}a--c), the angle $\beta$ can be written as
a function of $\theta$,
 \beqa
   \beta= -\arctan(\sin(\theta))
   \label{e:beta_theta}
 \eeqa
Finally, by substituting Eq.~(\ref{e:beta_theta}) into
Eq.~(\ref{equation:J}), the inverse Jacobian matrix $\negr J^{-1}$
can be simplified as follows
 \bed
 \negr  J^{-1}=
   \left[\begin {array}{ccc}
       1             &
       -\tan(\theta) &
       -\tan(\theta)\\
       -\tan(\theta) &
       1             &
       -\tan(\theta) \\
       -\tan(\theta) &
       -\tan(\theta) &
       1
       \end{array}
   \right]
 \eed
Thus, the square roots of the eigenvalues of $(\negr J \negr
J^T)^{-1}$ are,
 \bed
   \xi_1= |2 \tan(\theta)-1| \quad {\rm and} \quad
   \xi_2= \xi_3= |\tan(\theta)+1|
 \eed
And the three velocity transmission factors are,
 \be
   \psi_1= \frac{1}{|2 \tan(\theta)-1|}  \quad {\rm and} \quad
   \psi_2= \psi_3= \frac{1}{|\tan(\theta)+1|}
   \label{e:frontieres}
 \ee
Figure~\ref{figure:Variation_Xi} depicts $\psi_1$, $\psi_2$ and
$\psi_3$ as function of $\theta$ along the axis $(Q_1Q_2)$.

The joint limits on $\theta$ are located on both sides of the
isotropic configuration. To calculate the joint limits, we solve
the following inequations,
 \bseq
 \beqa
    \frac{1}{\psi_{max}} \leq \frac{1}{|2 \tan(\theta)-1|} \leq \psi_{max} \\
    \frac{1}{\psi_{max}} \leq \frac{1}{|\tan(\theta)+1|} \leq \psi_{max}
 \eeqa
 \label{e:constraints}
 \eseq
where the value of $\psi_{max}$ depends on the performance
requirements. Two sets of joint limits
($[\theta_{Q_1}~\beta_{Q_1}]$ and $[\theta_{Q_2}~\beta_{Q_2}]$)
are found. The detail of this calculation is given in the
Appendix.

The position vectors $\negr q_1$ and $\negr q_2$ of the points
$Q_1$ and $Q_2$, respectively, can be easily defined as a function
of $L$ (Figs.~\ref{figure:Workspace1} and
\ref{figure:Workspace2}),
 \bseq
 \beq
    \negr q_1= [q_1~q_1~q_1]^T
    \quad {\rm and} \quad
    \negr q_2= [q_2~q_2~q_2]^T
   \label{e:q1-q2}
 \eeq
with
 \beq
    q_1= - \sin(\beta_{Q_1}) L
    \quad {\rm and} \quad
    q_2= - \sin(\beta_{Q_2}) L
 \eeq
 \eseq

The size of the Cartesian workspace is,
 \bed
   L_{Workspace} = |q_2 - q_1|
 \eed
Thus, $L$ can be defined as a function of $L_{Workspace}$.
  \bed
    L= \frac{L_{Workspace}}{|\sin(\beta_{Q_2}) - \sin(\beta_{Q_1}) |}
  \eed

\begin{figure}[!ht]
    \begin{center}
           \includegraphics[width=72mm,height=68mm]{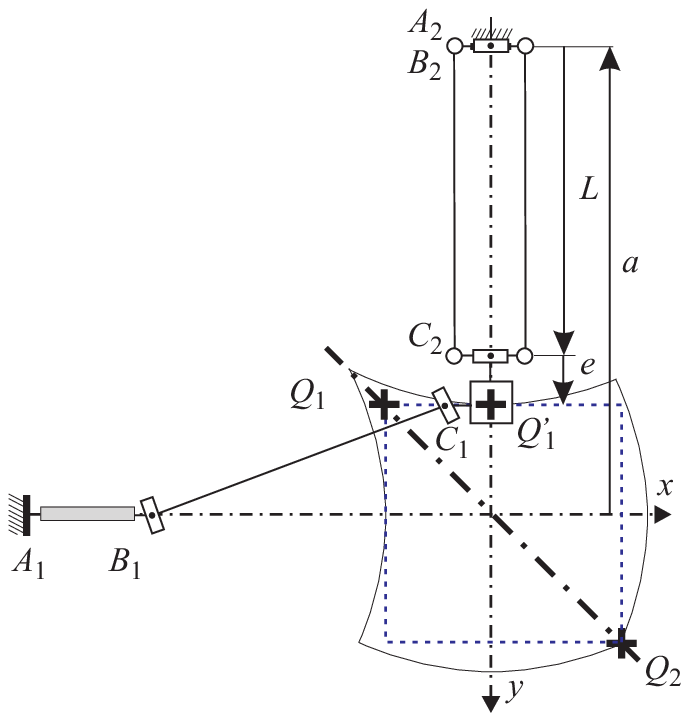}
           \caption{The point $Q_1'$ used for the determination of $a$}
           \protect\label{figure:Workspace3}
    \end{center}
\end{figure}

{\bf Step~3:} We want to determine the positions of the base
points, namely, $a$. When the tool center point P is at $Q_1'$
defined as the projection onto the $y$-axis of $Q_1$, $\rho=0$
and, (Fig.~\ref{figure:Workspace3})
\beqa
    OA_2=OQ_1'+Q_1'C_2+C_2A_2 \nonumber
\eeqa
 with $OA_2=a$, $OQ_1'=q_1$, $Q_1'C_2=PC_2=-e$ and since $\rho=0$, $C_2A_2=C_2B_2-L$.
 Thus,
 \bed
   a= q_1 -  e - L
 \eed

Since $q_1$ is known from Eqs.~(\ref{e:q1-q2}) and
~(\ref{e:joint-limit1}), $a$ can be calculated as function of $e$,
$L$ and $\psi_{max}$.

Now, we have to calculate the linear joint range $\Delta
\rho=\rho_{max}$ (we have posed $\rho_{min}$=0).

When the tool center point $P$ is at $Q_2$, $\rho=\rho_{max}$. The
equation of the direct kinematics (Eq.~(\ref{e:p-poistion}b))
written at $Q_2$ yields,
 \bed
    \rho_{max} = q_2 - a - \cos(\theta_{Q_2}) \cos(\beta_{Q_2}) L - e
 \eed
\subsection{Prototype}
\begin{figure}[!ht]
  \begin{center}
    \includegraphics[width=85mm,height=61mm]{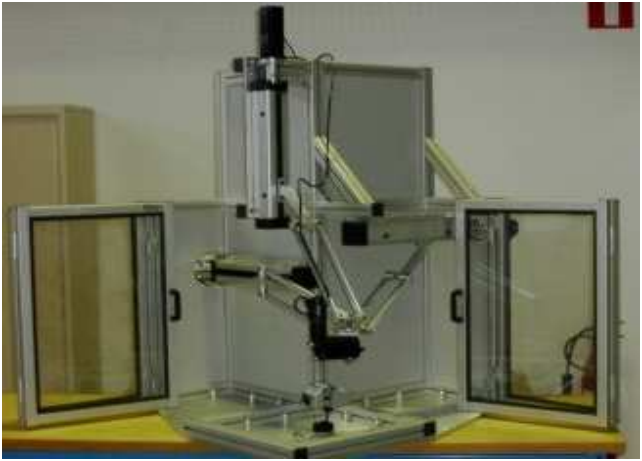}
    \caption{The orthoglide prototype}
    \protect\label{figure:Prototype}
  \end{center}
\end{figure}

\begin{figure}[!ht]
    \begin{center}
           \includegraphics[width=46mm,height=122mm]{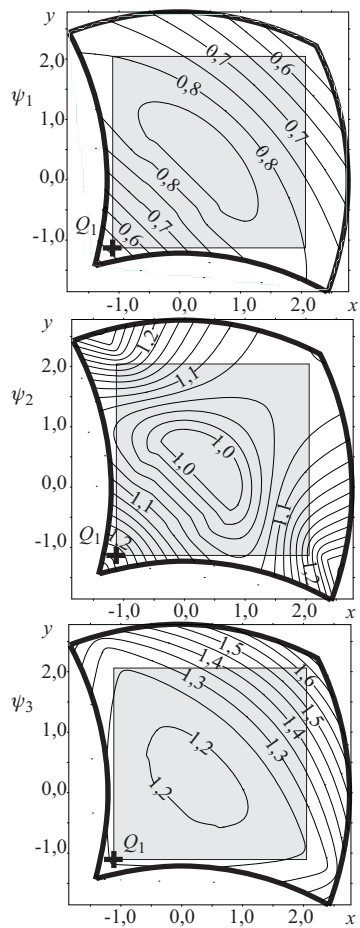}
           \caption{The three velocity transmission factors in a
           $z$-cross section of the Cartesian workspace passing through $Q_1$}
           \protect\label{figure:Isovalues_1}
    \end{center}
\end{figure}

Using the aforementioned two kinetostatic criteria, a small-scale
prototype is under development in our laboratory.
The mechanical structure is now finished,
(Fig.~\ref{figure:Prototype}).
The actuated joints used for this prototype are rotative motors
with ball screws. The prescribed performances of the orthoglide
prototype are a Cartesian velocity of $1.2 m/s$ and an
acceleration of $14m/s^2$ at the isotropic point. The desired
payload is $4 kg$. The size of its prescribed cubic workspace is
$200 \times 200 \times 200~mm$. We limit the variations of the
velocity transmission factors as,
 \be
   1/2 \leq \psi_i \leq 2
   \label{e:velocity_limits_prototype}
 \ee
The resulting length of the three parallelograms is $L=310~mm$ and
the resulting range of the linear joints is $\Delta~\rho=~
257~mm$. Thus, the ratio of the range of the actuated joints to
the size of the prescribed Cartesian workspace is
$r=200/257=0.78$. This ratio is high compared to other mechanisms.
The three velocity transmission factors are depicted in
Fig.~\ref{figure:Isovalues_1}. These factors are given in a
$z$-cross section of the Cartesian workspace passing through
$Q_1$.
\section{Conclusions}
Presented in this paper is a new kinematic structure of a PKM
dedicated to machining applications: the Orthoglide. The main
feature of this PKM design is its trade-off between the popular
serial PPP architecture with homogeneous performances and the
parallel kinematic architecture with good dynamic performances.

The workspace is simple, regular and free of singularities and
self-collisions. The Jacobian matrix is isotropic at a point close
to the center point of the workspace. Unlike most existing PKMs,
the workspace is fairly regular and the performances are
homogeneous in it. Thus, the entire workspace is really available
for tool paths. In addition, the orthoglide is rather compact
compared to most existing PKMs. A small-scale prototype of this
mechanism is under construction at IRCCyN. First
experiments with plastic parts will be conducted. The dynamic
analysis has not been reported in this article. A rigid dynamic model has been
proposed in (\cite{Guegan:2002} and an elastic dynamic model is now being
developed with the software package Meccano.
\section*{Acknowledgments}
The authors would like to acknowledge the financial support of
R\'egion Pays-de-Loire, Agence Nationale pour la Valorisation de
la Recherche, and \'Ecole des Mines de Nantes.

\section{Appendix}
To calculate the joint limits on $\theta$ and $\beta$, we solve
the followings inequations, from the Eqs.~\ref{e:constraints},
 \bseq
  \be
    |2\tan(\theta)-1| \leq \psi_{max}
  \ee
  \be
    \frac{1}{|2\tan(\theta)-1|} \leq \psi_{max}
  \ee
 \eseq
Thus, we note,
 \bseq
 \be
   f_1= |2\tan(\theta)-1| \quad
   f_2= 1/|2\tan(\theta)-1|
 \ee
 \eseq
\begin{figure}[!ht]
    \begin{center}
           \includegraphics[width=73mm,height=44mm]{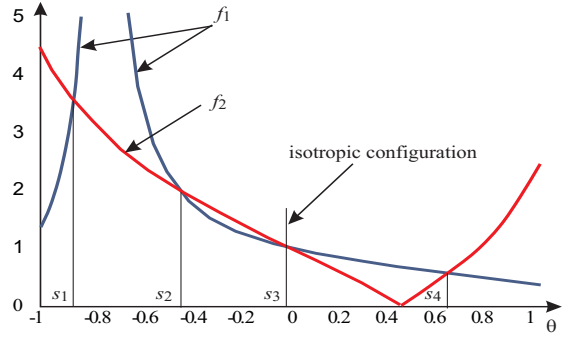}
           \caption{$f_1$ and $f_2$ as function of $\theta$ along $(Q_1Q_2)$}
           \protect\label{figure:variations_theta}
    \end{center}
\end{figure}
Figure~(\ref{figure:variations_theta}) shows $f_1$ and $f_2$ as
function of $\theta$ along $(Q_1Q_2)$. The four roots of $f_1=f_2$
in $[-\pi~\pi]$ are,
 \bseq
 \beqa
   s_1&=& -\arctan\left((1+\sqrt{17})/4\right) \\
   s_2&=& -\arctan\left(1/2\right) \\
   s_3&=& 0 \\
   s_4&=& \arctan\left((-1+\sqrt{17})/4\right)
 \eeqa
with
 \beqa
   f_1(s_1)= (-3+\sqrt{17})/4 \quad
   f_1(s_2)= 2\\
   f_1(s_3)= 1 \quad
   f_1(s_4)= (3+\sqrt{17})/4
 \eeqa
 \eseq
and
 \bseq
 \beqa
   f_1(\theta)=0 &{\rm when} & \theta= \arctan(1/2)-\pi \\
   f_2(\theta)=0 &{\rm when} & \theta= \arctan(1/2))
 \eeqa
 \eseq
The isotropic configuration is located at the configuration where
$\theta=\beta=0$. The limits on $\theta$ and $\beta$ are in the
vicinity of this configuration. Along the axis $(Q_1Q_2)$, the
angle $\theta$ is lower than $0$ when it is close to $Q_2$, and
greater than $0$ when it is close to $Q_1$.

To find $\theta_{Q_1}$, we study the functions $f_1$ and $f_2$
which are both decreasing on $[0~\arctan(1/2)]$. Thus, we have,
 \bseq
  \beqa
     \theta_{Q_1} &=&  \arctan\left({\frac {{\psi_{max}}-1}{{2\psi_{max}}}}\right)
     \\
     \beta_{Q_1}  &=& -\arctan\left({\frac {\psi_{max}-1}{\sqrt {5\psi_{max}^{2}-2\psi_{max}+1}}}\right)
     \label{e:joint-limit1}
  \eeqa
 \eseq
In the same way, to find $\theta_{Q_2}$, we study the functions
$f_1$ and $f_2$ on $[s_1~0]$. The three roots $s_1$, $s_2$ and
$s_3$ define two intervals. If $\psi_{max} \in
[f_1(s_1)~f_1(s_2)]$, we have,
 \bseq
  \beqa
     \theta_{Q_2} &=& -\arctan\left({\frac {{\psi_{max}}-1}{{\psi_{max}}}}\right)
     \label{e:joint-limit2} \\
     \beta_{Q_2}  &=&  \arctan\left({\frac {\psi_{max}-1}{\sqrt {2\psi_{max}^{2}-2\psi_{max}+1}}}\right)
  \eeqa
otherwise, if $\psi_{max} \in [f_1(s_2)~f_1(s_3)]$,
  \beqa
     \theta_{Q_2} &=& -\arctan\left({\frac {\psi_{max}-1}{{2}}}\right)
     \label{e:joint-limit3}  \\
     \beta_{Q_2}  &=&  \arctan\left({\frac {\psi_{max}-1}{\sqrt {{\psi_{max}}^{2}-2\psi_{max}+5}}}\right)
  \eeqa
 \eseq
\end{document}

%% file: macros.tex
\newcommand{\gbf}[1] {\mbox{\boldmath${#1}$\unboldmath}}
\newcommand{\be}{\begin{equation}}
\newcommand{\ee}{\end{equation}}
\newcommand{\beq}{\begin{equation}}
\newcommand{\eeq}{\end{equation}}
\newcommand{\bed}{\begin{displaymath}}
\newcommand{\eed}{\end{displaymath}}
\newcommand{\beqa}{\begin{eqnarray}}
\newcommand{\eeqa}{\end{eqnarray}}
\newcommand{\beqann}{\begin{eqnarray*}}
\newcommand{\eeqann}{\end{eqnarray*}}
\newcommand{\bseq}{\begin{subequations}}
\newcommand{\eseq}{\end{subequations}}

\newcommand{\ba}{\begin{array}}
\newcommand{\ea}{\end{array}}

\newcommand{\w}{{\bf w}}

\newcommand{\h}{{\bf h}}

\newcommand{\negr}[1]{{\bf {#1}}}